\pdfoutput=1

\documentclass[11pt]{article}

\usepackage[]{acl}
\usepackage{graphicx}
\usepackage{times}
\usepackage{amsmath}
\usepackage{latexsym}
\usepackage{booktabs}

\usepackage[T1]{fontenc}

\usepackage[utf8]{inputenc}
\usepackage{algpseudocode}
\usepackage{algorithm}
\usepackage{microtype}

\title{Fine-tuning a Subtle Parsing Distinction Using a Probabilistic Decision Tree: the Case of Postnominal
\emph{"that"} in  Noun Complement Clauses vs. Relative Clauses}

\author{Author 1 \and ... \and Author n \\
         Address line \\ ... \\ Address line}

\author{Zineddine Tighidet \\
  Université Paris Cité\\
   F-75013 Paris, France\\
  \textit{tighidet.zineddine@gmail.com} \\\And
  Nicolas Ballier \\
  CLILLAC-ARP and LLF, Université Paris Cité\\   \& CNRS  F-75013 Paris France\\
  \textit{nicolas.ballier@u-paris.fr} \\}

\begin{document}
\maketitle
\begin{abstract}
In this paper we investigated two different methods to parse relative and noun complement clauses in English and resorted to distinct tags for their corresponding \textit{that} as a relative pronoun and as a complementizer. We used an algorithm to relabel a corpus parsed with the GUM Treebank using Universal Dependency. Our second experiment consisted in using  TreeTagger, a Probabilistic Decision Tree, to learn the distinction  between the two complement and relative uses of postnominal \emph{“that”}. We investigated the effect of the training set size on TreeTagger accuracy and how representative the GUM Treebank files are for the two structures under scrutiny. We discussed some of the linguistic and structural tenets of the learnability of this distinction.
\end{abstract}

\section{Introduction}
English has relative clauses (\textit{the man that I saw}) and noun complement clauses (\textit{the fact that I saw a man}) that may have similar surface representations (often the definite article, a noun, often immediately followed by \textit{that}) but different structural properties 
\cite{ballier2004praxis}. For POS-tagging systems based on trigrams, the distinction between these constructions can be challenging, not to mention the case of ambiguous sentences such as "the suggestion that he was advancing was ridiculous" \cite{huddleston1984introduction}.This is an issue for information retrieval, as conceptual argumentation makes heavy uses of noun complement clauses \cite{ballier2007completive}, the governors of these noun complement clauses being "shell nouns" \cite{schmid2012english}. Complement taking nouns \cite{Bowen2005NounCI} are crucial for the expression of stance \cite{charles2007argument} in documents, which is why this distinction may matter more than is usually assumed.

The Penn Treebank \cite{marcus-etal-1993-building} tagset \cite{santorini1990part} does not make strict distinctions between the part-of-speech (POS) tag of \emph{"that"} when used as a relative pronoun (WDT) or when used as a conjunction when complementizing nouns: it uses IN when complementizing verbs or nouns. Even though the CLAWS8 \footnote{CLAWS, the Constituent Likelihood Automatic Word-tagging System, is the name of the tagset and of the POS-tagging software for English text, CLAWS \cite{garside1987claws}}. \cite{claws8} tagset encodes this distinction with the \emph{CST}\footnote{\emph{"that"} as a conjunction} and \emph{WPR}\footnote{\emph{"that"} as a relative pronoun} tags, this tagger is not free and remains the property of the \href{https://ucrel.lancs.ac.uk/}{University Centre for Computer Corpus Research on Language (UCREL)}. To the best of our knowledge, the precision and recall of these two tags (and their corresponding syntactic structures) have not been reported. 

Admitting POS-tagging systems have reached an overall satisfactory  precision rate for standard English tagsets, we claim that this is not necessarily the case for tags that reflect such a subtle distinction which may have very similar surface representations. Discussing such POS-tags involves parsing issues of the \emph{that}-clause that follows the noun. Our research question is mostly based on the ability of a system to identify noun complement clauses as apposed to (restrictive) relative clauses, but this can be addressed by analysing dependency relation labels (parsing) or distinct tags that encode this syntactic distinction (POS-tagging). We present the two strategies in two experiments, exploring whether such specific Universal Dependency labels can be learnt. In this paper, we only investigate overt complementizers as we are also investigating how \textit{that} is tagged and do take into account noun complement clauses with zero complementizer, like in the example \textit{Plus the fact I'm a coward} from the British National Corpus \cite{bnc2007british}.  

The rest of the paper is structured as follows. Section 2 details the data we used for our experiments. Section 3 analyses the Universal Dependency (UD) GUM Treebank for English in terms of precision for the dependency labels of these two structures as well as their distribution across the training, testing and development sets. We describe an experiment replicating one of the specific features of the GUM Treebank. Section 4 details an experiment based on algorithm adapting the UD annotation generated with GUM. Section 5 explains how Treetagger can be used to learn distinct tags for \textit{that} used as a relative pronoun (WPR) or as a complementizer (CST). Section 6 discusses our results and section 7 outlines our future research.

\section{Material and Methods}

\subsection{Test Sets}
For our validation procedure, we used two test sets NCCtest and RCtest, one including 194 noun complement clauses (NCC), the other one included 189 relative clauses (RC). As language is complex, some sentences included other syntactic realisations, and a couple of "distractors" representative of the alternate structure were therefore included in our two test sets. We specify in Table \ref{tab:stats} the expected (gold) label counts for each test set. Two annotators agreed on these gold labels of these two test sets ($\kappa=1$).

\subsection{Brown Corpus}
We used the Brown corpus \cite{francis1979brown}, which is rather small with its 1 M tokens by contemporary standards, but well-balanced and freely available. Its current distribution in the NLTK python library \cite{bird2006nltk} has been POS-tagged with the Penn Treebank, this is the substrate we used for our re-annotation experiment with TreeTagger. Treetagger is a probabilistic tagger which uses decision trees for probability transitions, which is robust for its retraining and claims accuracy above 96 \% \cite{schmid1994treetagger}.

\subsection{Universal Dependency Annotation with UDPipe}
UDPipe \cite{straka-2018-udpipe} is a pipeline that takes as input a text file and renders a CoNLL-U\footnote{\href{https://universaldependencies.org/format.html}{https://universaldependencies.org/format.html}} file which contains the language-specific part-of-speech tag (\emph{XPOS}), lemma or stem, the \emph{DEPREL} (universal dependency relation)  etc.\\
A file annotated in Universal Dependency contains among other columns the \emph{XPOS} (part of speech)  for each token and the dependency relation, \emph{acl:relcl} for relative clauses and (just) \emph{acl} for noun complement clauses, though this more general category (\emph{acl} corresponds to clausal modifier of noun, adnominal clause) also includes non-finite clause. 

\begin{figure}[!h]
\begin{center}
\includegraphics[scale=0.7]{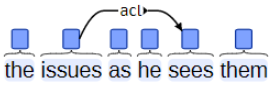} 
\caption{Example of clause modifier of noun (\emph{acl}).}
\label{fig.1}
\end{center}
\end{figure}

\textbf{Clausal modifier of noun (\emph{acl})} \\
\emph{acl} stands for finite and non-finite clauses that modify a noun. The governor (head) of the \emph{acl} dependency relation is the noun that is modified, and the dependent is the predicate  of the clause that modifies the noun. 
In Figure~\ref{fig.1} the finite clause “as he sees them” modifies the noun “the issues”.

As evidenced by this example taken from the UD documentation, \textit{acl} is a label that encompasses more than \textit{that} noun complement clauses.

\begin{figure}[!h]
\begin{center} 
\includegraphics[scale=0.7]{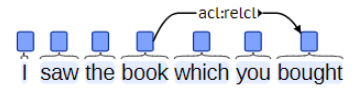} 

\caption{Example of relative clause modifier (\emph{acl:relcl}).}
\label{fig.2}
\end{center}
\end{figure}

\textbf{Relative clause modifier (\textit{acl:relcl})}\\
A relative clause modifier of a noun is a clause that modifies the antecedent. The \emph{acl:relcl} relation points from the governor (the antecedent) head of the modified nominal to the dependent (verb) of the relative clause. In Figure~\ref{fig.2} the relative clause “which you bought” modifies the nominal “the book”.

Several treebanks for English are available \footnote{\url{https://universaldependencies.org/treebanks/en-comparison.html}} for the Universal dependency annotation \cite{mcdonald2013universal}. We focused on the GUM Treebank \cite{Zeldes2017}, based on the Georgetown University Multilayer (GUM) Corpus \footnote{\url{https://gucorpling.org/gum/}} as its CoNLL-U format 
\footnote{an adaptation of the CoNLL-X format, see \cite{buchholz2006conll},
\url{https://universaldependencies.org/format.html}} contains a specific column that reports the dependency relation and the governor. Our next section analyses the accuracy of these two tags when labelling noun complement clauses and relative clauses in the development (DEV), training (TRAIN) and testing (TEST) sets of the GUM Treebank based on the GUM corpus \cite{gum}.



\section{Revisiting the GUM Treebank}
We noticed some debatable annotations for some cases where ellipsed \textit{such} and \textit{so} led some \textit{that}-clauses expressing consequence to be labelled as \textit{acl} as in \textit{"As a result, wikiHow is still at the size that every editor eventually gets to know other editors"}. We computed the proportion of Relative Clauses (RC) in relation to noun complement clauses (NCC).

\subsection{Frequency of RC and NCC in the GUM Treebank}
\begin{table}
\begin{center}

\begin{tabular}{ | c | c | c | c | } 
  \hline
  deprel & Train & Dev & Test \\ 
  \hline
  acl:that & 65 (0.513) & 13 (0.65) & 14 (0.69) \\ 
  \hline
  acl:relcl & 1419 & 258 & 216 \\ 
    &  (11.21) & (12.92) & (10.70) \\
  \hline
\end{tabular}
\caption{Frequency of "acl:relcl" and "acl:that" in the GUM Treebank files raw (normalized per 1000 tokens).}
\label{tab:freq_acl_GUM}
\end{center}
\end{table}
As can be seen in Table \ref{tab:freq_acl_GUM}, there are at least 15 times more relative clauses (RC) than noun complement clauses (NCC) in the GUM Treebank. 


One of the benefits of the GUM Treebank is that it contains extra information, the ninth column conflates the dependency relation (acl) and \textit{that} for noun complement clauses, we have tried to exploit this \textit{acl:that} tag by building a UDPipe model based on this treebank and by trying to recapture this information by an algorithm.

\subsection{Replicating the GUM Ninth Column}
In the ninth column of the GUM corpus, we were specifically interested in the \textit{"acl:relcl"} and \textit{"acl:that"} annotations to improve the detection of noun complement clauses, since the standard deprel (dependency relation) column only provides the \textit{"acl"} label and does not distinguish between finite and non finite uses of adnominal clauses. We trained a UDPipe model using the training, development and test sets of the  GUM Treebank on Github\footnote{\href{https://github.com/UniversalDependencies/UD_English-GUM}{https://github.com/UniversalDependencies/UD\_English\-GUM}}. However, once we applied the model on the same unannotated corpus, the ninth column was empty. It seems that UDPipe only captures the standard columns of the treebanks.
\subsection{Emulating the Ninth column}
We were therefore interested in reconstructing this column by implementing a heuristic. Once the acl:relcl have been copied from the deprel column, the algorithm consists in exploiting the seventh (Head of the current word) and eighth (Universal dependency relation to the HEAD) columns such that:

\algnewcommand\algorithmicforeach{\textbf{for each}}
\algdef{S}[FOR]{ForEach}[1]{\algorithmicforeach\ #1\ \algorithmicdo}

\begin{algorithm}
\caption{: \textbf{Heuristic to emulate acl:that labels in the ninth column}}
\begin{algorithmic}
\ForEach{$sentence \in corpus $}
\ForEach {$token \in sentence $}{ \\
\begin{enumerate}
    \item{Combine the seventh and eighth columns of the $token$ that were generated by the previously trained UDPipe model.}
    
    \item{If \textit{"that"} is right after the word to which the seventh column of the $token$ points to, then add \textit{"that"} to the ninth column.}
\end{enumerate}
}
\EndFor
\EndFor
\end{algorithmic}
\end{algorithm}
\section{Learning to tag with TreeTagger}
This retagging experiment \cite{inbook_gaillat_al} relies on the ability of TreeTagger \cite{schmid1994treetagger} to be used not only as a POS-tagger but as a tool which can be trained to learn how to tag, provided a specific tagset and sample data are provided. We used samples from the Brown corpus in its NLTK distribution and modified the Penn Treebank tagset to distinguish \textit{that} as WPR (relative pronoun) and  \textit{that} as CST (complementizer). In the learning phase, TreeTagger sees a vocabulary file and tokens associated to their tags and generates a .par model file to be used for POS-tagging. 
This section describes how we modified the tags to train the system \footnote{The Python implementation is available in this GitHub repository: \href{https://github.com/Zineddine-Tighidet/Relative-Complement-That-Annotator}{https://github.com/Zineddine-Tighidet/Relative-Complement-That-Annotator}}. After the annotation of the Brown corpus by UDPipe, a heuristic was applied on the results in order to introduce the \emph{WPR} and \emph{CST} tags which are not previously used in the tagset. To do that, the \emph{DEPREL} label was used, so our method assumes that the UDPipe trained with the English GUM corpus provides a sufficiently correct \emph{DEPREL} label for noun complement clauses:

\algdef{S}[FOR]{ForEach}[1]{\algorithmicforeach\ #1\ \algorithmicdo}

\begin{algorithm} \caption{: \textbf{Heuristic for Brown re-annotation}}
\begin{algorithmic}
\ForEach{$sentence \in corpus $}
\ForEach {$token \in sentence $}{ \\
\begin{itemize}
    \item{If the $token$ is a verb (i.e. XPOS = VB) and is a clausal modifier of noun (i.e. DEPREL = acl) then go steps before that $token$ to see if there is any \emph{"that"}, if so, label it as \emph{CST}.}
    
    \item{If the $token$ is a verb (i.e. XPOS = VB) and is part of a relative clause (i.e. DEPREL = acl:relcl) then go steps before that $token$ to see if there is any \emph{"that"}, if so, label it as \emph{WPR}.}
\end{itemize}
}
\EndFor
\EndFor
\end{algorithmic}
\end{algorithm}

The aim of this experiment is to see how the TreeTagger accuracy increases as a function of the training set size. To do this, the TreeTagger received different proportions of a training set as input. To be more specific, there are 500 training files representing the annotated Brown corpus, for the first training the first 10 files were used, and then the 30, 100, 200, 300, 400 and finally the 500 training files. For each training a \emph{.par} file that corresponds to the model was returned.

\section{Results}

\subsection{Emulating the Ninth column}
To assess this algorithm that selects only \textit{that}- (finite) clauses among the acl clauses, we tested it with the GUM treebank, comparing our results in our reconstructed column with the original data. The heuristic gave good results for the annotations of relative clauses  "acl:relcl" with an accuracy that exceeds 90\% (see table \ref{tab:acc_table_aclrelclthat}).


\begin{table}
\begin{center}
\begin{tabular}{ | c | c | c | c |} 
  \hline
  deps column & Train & Dev & Test \\ 
  \hline
  acl:that & 0.78 & 0.76 & 0.71 \\ 
  \hline
  acl:relcl & 0.92 & 0.92 & 0.94 \\
  \hline
\end{tabular}
\caption{Accuracy of acl:relcl and acl:that annotations in the "deps" column recreated by combining the "head" and "deprel" columns for each of the GUM Treebank files.}
\label{tab:acc_table_aclrelclthat}
\end{center}
\end{table}
Nevertheless, the algorithm works less well  for "acl:that", this is partly due to some coordinated NCC clauses and to multi-word-units (like \textit{quid pro quo}).

\subsection{Re-annotating with TreeTagger}
\begin{figure}[!h]
\begin{center}
\includegraphics[scale=0.5]{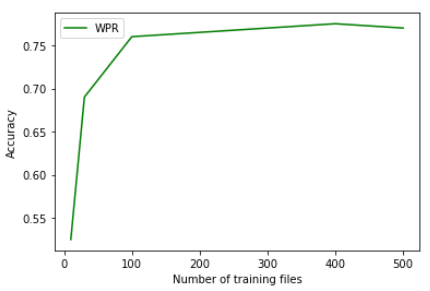} 

\caption{TreeTagger accuracy curve for \emph{WPR} tag (computed using the RCtest data).}
\label{wpr_fig}
\end{center}
\end{figure}

\begin{figure}[!h]
\begin{center}
\includegraphics[scale=0.5]{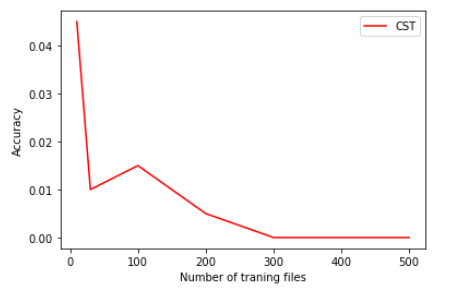} 
\caption{TreeTagger accuracy curve for \emph{CST} tag (computed on the NCCtest data).}
\label{cst_fig}
\end{center}
\end{figure}

We used our specifically designed testing files that contain respectively 
189 \emph{"that"} as \emph{WPR} and 194 \emph{"that"} as \emph{CST}. The first one named RCtest (Relative Clause) was used to compute the accuracy for \emph{WPR} and the second one named NCCtest (Noun Complement Clause) for \emph{CST} (see Figure ~\ref{wpr_fig} and ~\ref{cst_fig}). We used these specific files because they are manually annotated and each one of them contains a majority of the two tags we are interested in, which makes it convenient for our experiments.\\

As shown in Figures ~\ref{wpr_fig} and ~\ref{cst_fig} the TreeTagger accuracy increases with the number of training files for the RCtest data. This is a natural behaviour from a probabilistic model such as the TreeTagger, the probability increases as the weight of Relative clauses increases in the data it has. However, there is a drastic decrease in the \emph{CST} curve around 100 training files, and the TreeTagger did not perform very well annotating the \emph{"that"} with \emph{CST} tag, as shown with Figure~\ref{cst_fig} the accuracy is very low, and as the number of training files increases, the accuracy goes down. Whatever technique is used, the detection of noun complement clauses is more challenging than for relative clauses.

\section{Discussion} 
Two approaches for explaining the results obtained for the CST tag can be either statistically or linguistically motivated. Starting with the first approach, as shown in Table \ref{tab:stats}, as the number of training files goes up, the number of other tags increases, especially for the IN tag, which in this case represents a confusion by the TreeTagger to annotate with the right tag, in fact IN is not a specific tag but rather a generic tag as it also corresponds to verbal \textit{that}-clauses, therefore, this shows that the TreeTagger generated noise due to a confusion on the annotation of "that" (this is better illustrated in the figure \ref{vocab_compet} especially for the graph that represents the IN tag in blue.) The second approach consists in analysing the competing labels for \textit{that}.
\subsection{Accuracy in relation to other categories}
The Penn Treebank tagset \cite{santorini1990part}, even though it does not acknowledge the whole complex range of functional realisations of \textit{that}, e.g. adverbial, proform vs deictic uses, see \cite{Ballier2022} can help visualise the complex interaction of the learning process of the identification of the different functional uses of \textit{that}. As the training data increases, the variable proportions of the different functional realisations of \textit{that} probably changes, so that a probabilistic tagger generates models variable in their results for this tagging task. The tagger has to learn the different competing tags for \textit{that}. Our two test datasets allow us to monitor the evolution of the training phase as the size of the training data increases. Whereas we tried to train Treetagger to learn CST for NCC \textit{that} and WPR for relative pronouns, we also computed the distribution of other tags that "\textit{that}" may take, such as "WDT" (\textit{that} when used as a relative pronoun, but also "WH"-determiners such as \textit{which}), "DT" (Determiners), and "IN" (Subordinating conjunction, whether for nouns or for verbs) for each of the RC and NCC corpus. Table \ref{tab:stats} recaps the changes observed when we evaluated the labels with our two testing sets (RCtest and NCCtest). For each testing set, we indicate the expected count of each label in the columns RCtest GOLD and NCCtest GOLD.

\begin{table}[!h]
\centering
\resizebox*{8cm}{0.72\textheight}{%
\begin{tabular}{@{}lcccc@{}}
\toprule
     & \textbf{RCtest} & \textbf{RCtest GOLD} & \textbf{NCCtest}  & \textbf{NCCtest GOLD} \\ 
\midrule
 \textbf{\small{10 training files}}\\
 \hline
    \emph{WPR}   & 107  & 189 & 20    & 17    \\
    \emph{CST}   & 22  & 26  & 10     & 194     \\
    \emph{IN}  & 95  & 0 & 183   & 0       \\
    \emph{DT}  & 7  & 15  & 16     & 14     \\
  \midrule
     \textbf{\small{30 training files}}\\
     \hline
    \emph{WPR}   & 146 & 189  & 28    & 17    \\
    \emph{CST}   & 5  & 26  & 3     & 194     \\
    \emph{IN}  & 72  & 0 & 189   & 0       \\
    \emph{DT}  & 8   & 15 & 9     & 14     \\
  \midrule
   \textbf{\small{100 training files}}\\
   \hline
    \emph{WPR}   & 158 & 189  & 25    & 17    \\
    \emph{CST}   & 2  & 26  & 3     & 194     \\
    \emph{IN}  & 65 & 0  & 194   & 0       \\
    \emph{DT}  & 6  & 15  & 7     & 14     \\
    \midrule
   \textbf{\small{200 training files}}\\
   \hline
    \emph{WPR}   & 156 & 189  & 27    & 17    \\
    \emph{CST}   & 1  & 26  & 1     & 194     \\
    \emph{IN}  & 66  & 0 & 196   & 0       \\
    \emph{DT}  & 8  &  15 & 0     & 14     \\
    \midrule
   \textbf{\small{300 training files}}\\
   \hline
    \emph{WPR}   & 157  & 189 & 22    & 17    \\
    \emph{CST}   & 2  & 26  & 0     & 194     \\
    \emph{IN}  & 67 & 0  & 202   & 0       \\
    \emph{DT}  & 5  & 15  & 5     & 14     \\
    \midrule
   \textbf{\small{400 training files}}\\
   \hline
    \emph{WPR}   & 159 & 189 & 21    & 17    \\
    \emph{CST}   & 2 & 26   & 0     & 194     \\
    \emph{IN}  & 64 & 0  & 199   & 0       \\
    \emph{DT}  & 6  & 15  & 7     & 14     \\
    \midrule
       \textbf{\small{500 training files}}\\
   \hline
    \emph{WPR}   & 158 & 189  & 23    & 17    \\
    \emph{CST}   & 1  & 26  & 4     & 194     \\
    \emph{IN}  & 65  & 0 & 188   & 0       \\
    \emph{DT}  & 7  &  15 & 7     & 14     \\
  \bottomrule                          
\end{tabular}%
}
\caption{Statistics about \emph{WPR}, \emph{CST}, \emph{IN} and \emph{DT} tags obtained for each of the 7 models (i.e. trained with 10, 30, 100, 200, 300, 400 and 500 files).}
\label{tab:stats}
\end{table}

\begin{figure}[!h]
\begin{center}
\includegraphics[scale=0.5]{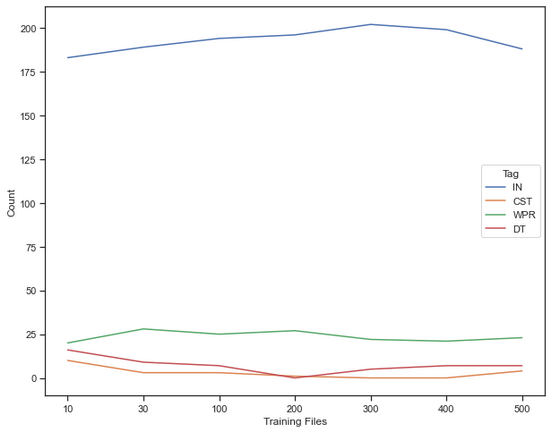} 

\caption{Evolution of IN, CST, WPR and DT tags with training files in the NCCtest corpus.}
\label{count_evol_ncctest}
\end{center}
\end{figure}

\begin{figure}[!h]
\begin{center}
\includegraphics[scale=0.5]{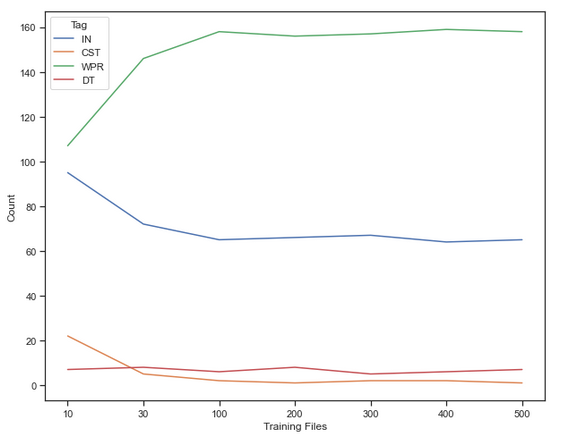} 

\caption{Evolution of IN, CST, WPR and DT tags with training files in the RCtest corpus.}
\label{count_evol_rctest}
\end{center}
\end{figure}

Here is an example of these potential mishaps in the POS-tagging: "that meeting that|IN [vs DT] morning was about a public case that|IN [vs WPR] we might make". The first deictic \textit{that} was properly labelled, the second one was erroneously labelled as a subordinating conjunction and for the third occurrence, the relative pronoun was tagged as a subordinating conjunction (see additional examples in the Appendix).  

\subsection{Weakness of the TreeTagger-based heuristic} \label{piste}

We re-annotated a corpus initially tagged with the Penn Treebank, which means that we modified some IN tags to CST and some IN tags to WPR for relative pronouns but the Brown corpus data retained some WDT labels. As shown in Table~\ref{tab:stats}, there are many \emph{WDT} tags, this is simply because the \emph{WDT} tag is both an older and more general version of the \emph{WPR} tag, and seemingly the TreeTagger kept the older version. So the \emph{WDT} and \emph{WPR} tags are likely labels for relative pronouns considered as equivalent in the computing of the metrics, even though strictly speaking some WDT tokens in the Brown corpus may correspond to WH-determiners such as \textit{which}.
The main objection to our method is that we only relabelled a portion of the IN tags, so that the system has to learn a WPR versus CST distinction while still being fed with some examples of IN. In this sense, we can only partially monitor the behaviour of Treetagger when subjected to more examples. Figure 
\ref{count_evol_ncctest} and Figure \ref{count_evol_rctest} plot the evolution of the tagging of the NCCtest and RCtest sets (respectively) as the corpus size increases. We expect the system to learn to relabel IN as either WPR or CST but this is hardly the case for CST. It should be noted that we did not control the input of the respective number of examples with CST and with WPR when increasing the data size of the training data. We only report the total counts of the tags assigned to \textit{that}, we did monitor the individual behaviour of the tagging system for each occurrence of \textit{that}.


\subsection{Long-Distance Dependencies} \label{explor}
\begin{figure}[!h]
\begin{center}
\includegraphics[scale=0.5]{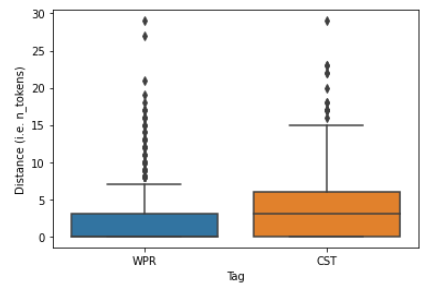} 
\caption{Distribution of the distance (number of tokens) separating a \emph{"that"} and the last noun before it for each of the \emph{WPR} and \emph{CST} \emph{"that"}.}
\label{boxplot}
\end{center}
\end{figure}

As already pointed out, noun complement clauses can follow a relative clause for the same noun (but not the other way round). \textit{That}-relative clauses tend to be adjacent to their antecedents (and are often restrictive relative clauses) whereas (\textit{that-}) noun complement clauses can be separated from their governor. So we explored a simple metric which is the distance (i.e. number of tokens) separating a \emph{"that"} (annotated either with \emph{CST} or \emph{WPR}) and the last noun before it. As shown in the boxplots in Figure~\ref{boxplot} there is a tendency showing that the \emph{"that"} tagged with \emph{CST} using a verb with a \emph{DEPREL} = \emph{acl} have a higher distance separating them from the last noun before them. This can probably cause some ambiguity due to the higher distance. However, as we can see for the \emph{"that"} tagged with \emph{WPR} using a verb with a \emph{DEPREL} = \emph{acl:relcl} the distance with the last noun is smaller, and there are less misclassifications (i.e. less noise) for the \emph{"that"} used as \emph{WPR}.
This is just a statistical approach to see if there is any bias that can explain why the heuristic produces a lot of noise. 

Our metric is rather crude but head nouns of NCCs need not be adjacent to the \textit{that}-clauses, so that an inventory of structures in-between could be taken into account. The distance between the governor and the \textit{that}-clause of these long-distance dependencies \cite{osborne2019dependency}
could be more systematically investigated. 

\subsection{Relevance of UD Deprel Labels for NCC?}
It should be noted that UD changed the dependency label for noun complement clauses, as explained on the UD website: "In earlier versions of SD/USD, complement clauses with nouns like \textit{fact} or \textit{report} were also analyzed as ccomp [clausal complement]. However, we now analyze them as acl. Hence, ccomp does not appear in nominals. This makes sense, since nominals normally do not take core arguments." We may challenge this view since ccomp implies a "clausal complement" and nouns may require a "core argument", even more so than for adjectives.\footnote{For a similar argumentation see \cite{osborne2019status}.}
One of the unfortunate consequences is that adverbs like \textit{now} in the sentence "Now that the world is in the age where lighting seems to be a daily necessity" are labelled as a governor of the "adnominal" clause. 
It maybe the case that acl is a debatable label, also used after verbs as for \textit{that} verbal complement clauses ("if this seems incredibly far-fetched, comfort yourself that double chute failure in modern times is also extremely unlikely, and that you have already beaten worse odds"). Consequently, the (SUD) Surface-Syntactic Universal Dependencies \cite{gerdes2018sud} has suggested alternative labels for acl. Another approach might be to restrict noun complement clauses to a subcategory of acl specific to noun complement clauses (possibly labelled as acl:ncl). 

\section{Further Research}
\subsection{Quality Monitoring of the Training Phase}
We have only estimated the accuracy of the annotation on our testing sets but we have not monitored the qualitative aspect of the annotation. Are some sentences systematically mislabelled or can we observe some changes during the training phase? For example, this NCC gets to be interpreted as a relative clause: \textit{ “O'Neill had an emotional reaction that [tagged as WPR] the level of corruption was too high to do serious projects in Russia,” Deripaska recalls.} Some configurations seem to remain challenging for parsing, and qualitative monitoring of the accuracy should take into account these sentences for which labelling improves or not. Controlling for frequency of exposure in the training data should prove to be very fruitful to maybe detect thresholds in  frequency (or proportions) in the training data for accurate tagging. For example, an example in our appendix seems to suggest that a trigram sequence \textit{no/N/that} (and corresponding identification of noun complement clauses) seems to be learned after exposure to the 100 training files (36 occurrences). As some of the examples of mislabellings in the Appendix also suggest, it is likely that our relabelling algorithm for WPR is too greedy, and a more elaborate version should filter out alternative relative pronouns that should inhibit the relabelling process. We should also apply stricter conditions on the type of \textit{that} which can be re-tagged. Assuming the DT label is correct, only IN labels should be re-tagged.


\subsection{More data?}
\begin{figure}[!h]
\begin{center}
\includegraphics[scale=0.5]{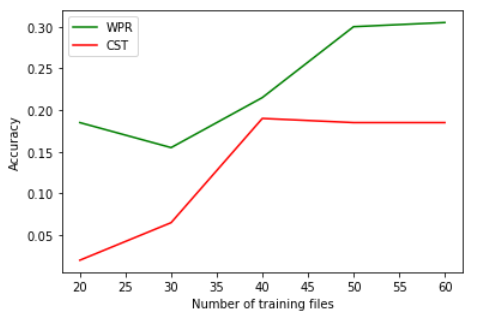} 

\caption{TreeTagger accuracy for \emph{"that"} annotated with \emph{CST} in red and \emph{WPR} in green with more training files.}
\label{more_data}
\end{center}
\end{figure}

More training files from the Brown corpus have been manually annotated and given to the TreeTagger, and an improvement in the \emph{CST} accuracy was observed (see Figure~\ref{more_data}). Though a plateau seems to be observed for the tag CST (\textit{that} for NCC complementizer), one may wonder if more examples of NCCs in the training data would alter this curve.   
We have only analysed the GUM Treebank for the UD analysis, but no less than six treebanks are available on github for the Universal Dependency analysis of English.

\subsection{Learnability and Dispersion}
\begin{figure*}
\begin{center}
\includegraphics[scale=0.35]{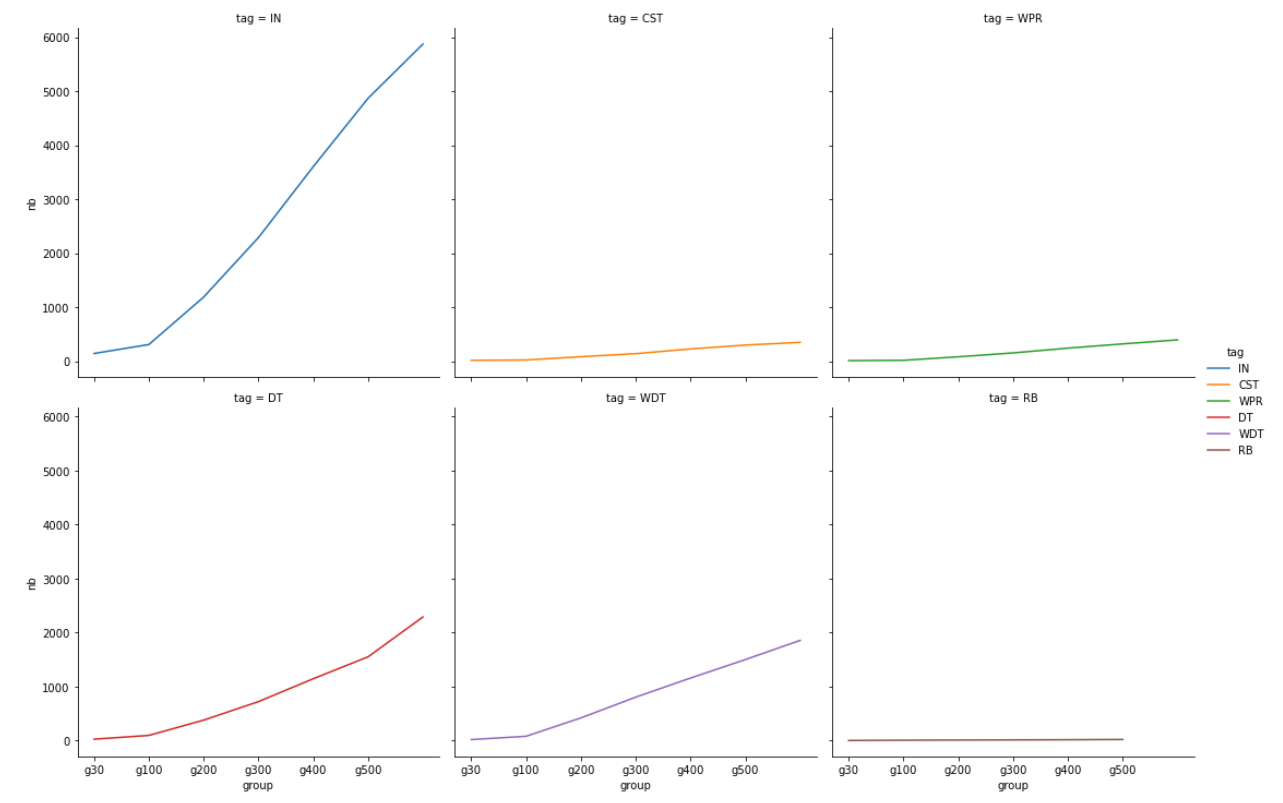} 
\caption{Evolution of the number of different tags for the re-annotated Brown corpus file groups (10, 30, 100, 200, 300, 400, 500 files.)}
\label{vocab_compet}
\end{center}
\end{figure*}

Our monitoring of the learning curve of the tag distinction in our TreeTagger experiment could be finer-grained: we did not control for genre types within the Brown corpus and the relative distributions of the two structures. If relative clauses seem to be more frequent than NCCs in the GUM Treebank, NCCs are more likely to be more frequent in argumentative texts \cite{ballier2007completive}. Our experiment only reported the effect of the number of the Brown files in the training data, not the specific distribution of the two structures across the different registers of the Brown corpus. The dispersion of these linguistic structures in the training data could be monitored across the corpus subparts using adequate dispersion measures \cite{Gries2020} or by comparing the vocabulary growth curves \cite{evert2007zipfr} of the two constructions across the Brown corpus files. Our Figure \ref{vocab_compet} crudely plots the distribution of the different tags in the training data as the size of the corpus increases (measured in number of files, but not with the corresponding text genres). Increasing the size of the corpus may require more attention to a frequency/textual diversity trade off.



\section{Conclusion}
In this study, we have experimented two methods to detect noun complement clauses, either by using the universal dependency GUM treebank or by retagging  the Brown corpus with specific \emph{WPR} and \emph{CST} tags. We also explored an automated way to do this annotation using a specific heuristic. We have evidenced the longer distance between the noun and the \textit{that}-clause for noun complement clauses.
The detection of relative clauses does seem to be much more robust than for noun complement clauses, which remains a problem for information retrieval as text genres could be interestingly classified with this criterion. The difference in frequency and in adjacency may account for such a discrepancy in the accuracy of the identification of the clause type. We have only begun to explore the parameters of the learnability of these tags corresponding to such a subtle linguistic distinction.

\section*{Acknowledgements}
We thank the three reviewers for their careful comments on a preliminary version of this paper. Thanks are due to Université Paris Cit{é} MSc in  Machine Learning for Data Science, which triggered this joint paper. Part of this research was carrried out on a CNRS research leave at the Laboratoire de Linguistique Formelle CNRS research lab, for which grateful thanks are acknowledged. We thank Issa Kant{é} for his collection of examples for the test datasets, partially reflecting his PhD data \cite{kante2017etude}.

\section*{Appendix}

\subsection{Example of a noun complement clause where \textit{that} gets properly tagged after 100 files in the training data (containing 36 occurrences of the \textit{no N that} sequence)}

\textit{"However, there is no guarantee \textbf{that}[tagged as CST] only the genuine repentant will produce works of value to the society."}

\subsection{Examples of remaining errors in our test sets}

We include examples of persistent mislabelling in our test data. After 500 training files, 6 sentences with \textit{that} in noun complement clauses are still tagged as if they were relative pronouns (with WPR).


\begin{itemize}
    \item \emph{The statement that|WPR the tribunal has made an "error of law" means no more or less than that|CST the construction placed upon the term by the court is preferred to that|DT of the tribunal.}

    \item \emph{There was no dispute that|WPR Bunn throughout acted with the authority of the bank.}

    \item \emph{This included a commitment that|WPR “if one of the two states should become the target of aggression, then the other side will give the aggressor no military aid or other support”.}

    \item \emph{We have received information that|WPR today, between 1400 and 1500, there was an explosion at the residence of Seyed Ali Khamenei.}

    \item \emph{Recently there was the illusion that|WPR Hamas, while not a perfect partner, was at least a group that could implement decisions,” he said.}
       
    \item \emph{Where there is a contract for the sale of goods by description, there is an implied condition \textbf{that}|WPR  the goods shall correspond with the description.}
\end{itemize}

\subsubsection{Example from our test sets that has been annotated with DT rather than with CST}

We illustrate the complexity of the polyfunctionality of \textit{"that"} by showing an example of overfitting for the deictic/pronominal uses of  \textit{"that"}.

\textit{"A high-ranking official in the Clinton administration expressed shock \textbf{that}[tagged as DT rather than CST] “the kids” in the White House “did not stand up when the president entered the room."}

\subsubsection{Examples from the RC test set that have been annotated with IN rather than with WPR}

\begin{itemize}
    \item \emph{High death rates among children reduce the value \textbf{that} |IN parents place on education; and so on.}
    \item \emph{The distinction \textbf{that} |IN matters is from that of 'patronage', which itself, as we shall see, is highly varied.}
\end{itemize}

\subsubsection{Examples from the NCCtest set that have been annotated with IN rather than with CST}

\begin{itemize}
    \item \emph{They're living proof \textbf{that} asthma can be passed from generation to generation.}
    \item \emph{Where there is a contract for the sale of goods by description, there is an implied condition \textbf{that} the goods shall correspond with the description.}
\end{itemize}

\subsection{An example of false positives for the Brown relabelling heuristic} \label{annot}

\begin{itemize}
    \item{\emph{"... But one does not have to affirm the existence of an evil order irredeemable in \textbf{that}[tagged as WPR] sense, or a static order in which no changes will take place in time, to be able truthfully to affirm the following fact: there has never been justitia imprinted in social institutions and social relationships except in the context of some pax-ordo preserved by clothed or naked force ..."} (it should be \emph{DT} rather than \emph{WPR}). The relative clause is with WHICH, not with THAT.}

\end{itemize}






\bibliography{anthology,custom}

\end{document}